# Visual Place Recognition with Probabilistic Voting

Mathias Gehrig, Elena Stumm, Timo Hinzmann and Roland Siegwart
*Autonomous Systems Lab, ETH Zurich*

*Abstract*— We propose a novel scoring concept for visual place recognition based on nearest neighbor descriptor voting and demonstrate how the algorithm naturally emerges from the problem formulation. Based on the observation that the number of votes for matching places can be evaluated using a binomial distribution model, loop closures can be detected with high precision. By casting the problem into a probabilistic framework, we not only remove the need for commonly employed heuristic parameters but also provide a powerful score to classify matching and non-matching places.
We present methods for both a 2D-2D image matching and a 2D-3D landmark matching based on the above scoring. The approach maintains accuracy while being efficient enough for online application through the use of compact (low-dimensional) descriptors and fast nearest neighbor retrieval techniques. The proposed methods are evaluated on several challenging datasets in varied environments, showing state-of-the-art results with high precision and high recall.

## I. INTRODUCTION AND RELATED WORK

Efficient and robust place recognition is of paramount importance for localization and mapping (SLAM) systems that seek accurate localization and drift-free maps. On a mission to explore unknown places, most robots construct a map by incrementally inferring their position from sensor data. In many situations, absolute position measurements are not available, resulting in an accumulation of drift over time. This issue can be addressed by solving two problems simultaneously: Firstly, detecting if the current place has been visited before, and secondly, associating the current place with the set of data that represents the revisited location.

The resulting task, typically referred to as place recognition or loop closure, is frequently solved on the basis of appearance due to the almost universal presence of cameras on mobile platforms and the rich information they provide [1]. In addition, in the context of SLAM, visual cues are often structured in the form of pose-graphs of visual landmarks and their observers [2]. The goal of this work is therefore to provide an improved framework for visual place recognition, relying on these sparse local features as input.

Much of the related work that strives for real-time place recognition discretizes the descriptor space to represent places in an efficient manner. Sivic *et al.* [3] introduced the popular bag-of-words (BoW) model. Here, classes are formed in descriptor space (*e.g.* using k-means clustering), to which descriptors extracted from an image are assigned. This way, a sparse binary feature vector with the dimension of the number of classes is built and used to efficiently represent the observed scene. However, the descriptor space discretization step can lead to deteriorated performance if the trained vocabulary does not represent the descriptor space accurately. Nonetheless, extensive research in this area has brought forward several well-performing algorithms such as [4, 5, 6]. For example, the work of Cummins and Newman [4] improves robustness by introducing a well-grounded probabilistic formulation of the problem.

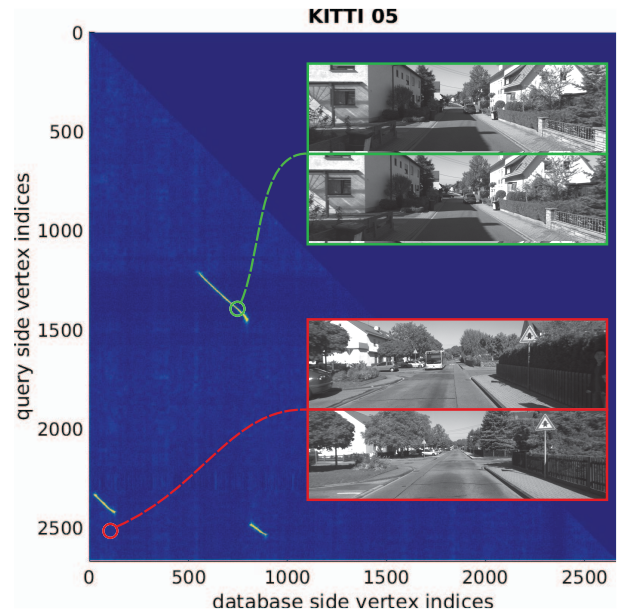

Fig. 1. Probabilistic scoring allows for a confident match of the same place (top) while rejecting non-matching places (bottom). The bright yellow stripes indicate high confidence that a place has been revisited. In this dataset, the car revisits three distinct sequences of the trajectory that allow for loop closures.

Discretization of descriptor space and BoW retrieval can be interpreted as an approximate nearest neighbor search [7]. Many recent approaches bypass descriptor quantization for potentially more accurate nearest neighbor searches. For example, Schindler et al. [8] performed approximate nearest neighbor search with vocabulary trees to efficiently retrieve the best matching image in the database by aggregating votes per image. In a similar fashion, Cieslewski *et al.* [9] used a kNN (k-nearest neighbors) voting scheme to retrieve the best matching keyframe to find loop closure candidates. While the former work is concerned with the localization problem, the latter focuses on loop closure detection. Loop closure detection algorithms have the added challenge of deciding whether a place has been revisited or not. This classification usually hinges on a parameter which is, in particular for kNN voting schemes, difficult to design. In an attempt to devise this parameter, Cieslewski et al. [9] normalize the score of a database[1] image with the number of landmarks it observes. However, this normalized score is unintuitive and still not independent of the number of nearest neighbors returned for a given query image or the number of descriptors in the database. Another approach is proposed by Lynen et al. [10], which also uses a kNN search and matches places by extracting regions of high vote density in a 2-D space of descriptor votes. In this case, the mean of the vote density in the candidate region was used as the decision making parameter. In contrast to Lynen's work, this paper is not dedicated to evaluating different place formulations, but to finding a method for improved reasoning within a descriptor voting framework.

More specifically, this paper presents a probabilistic approach to the problem of scoring based on aggregated descriptor votes. The method is not only efficient, but also offers state of the art performance by only considering single images as places. Efficient nearest neighbor search techniques directly retrieve descriptor matches, which then vote for images in the database. Instead of naively scoring the images by the number of votes or introducing heuristic normalization, we show that a probabilistic score based on the binomial distribution can be derived. Apart from providing a higher level of intuition, this score can be used to reliably classify loop closures, even in presence of strong perceptual aliasing as illustrated in figure 1. In addition, we demonstrate how co-visibility information can be combined with the probabilistic score to extract a set of landmark matches that represent the current place accurately.

This paper offers the following contributions:
- A novel probabilistic scoring method based on aggregated descriptor votes, which is not biased by the number of descriptors in images.
- Two resulting loop closure detection methods:
  ○ an algorithm that matches pose-graph vertices (vertex-to-vertex),
  ○ and an extension thereof that matches to the corresponding set of landmarks in the map (vertex-to-map).
- Efficient implementation methods which can speed up computation in the case of large pose-graphs.
- Quantitative evaluation and discussion of the presented approaches on three substantially different test environments.

## II. METHODOLOGY

*A. Terminology*

The proposed algorithm runs on a sparse feature map created by a pose-graph SLAM algorithm [11]. A map consists of a set of landmarks which are points in three-dimensional space with associated descriptors. Each node (or vertex) in the pose-graph is associated with at least one image. In case of multi-camera systems, the images are grouped together in a vertex such that votes are cast on vertices instead of images.

*B. Overview*

The core algorithm of this system is based on direct nearest neighbor search in projected descriptor space as Lynen *et al.* [10] proposed it for loop closure detection. After aggregating matches in vertices of the pose-graph, a binomial distribution is used to classify matching and non-matching places efficiently.

*C. Descriptor Projection*

In order to accelerate approximate nearest neighbor search, a 384-bit version of BRISK [12] is projected into a lower dimensional, real-valued space. As proposed in [10], the target dimensionality was set to 10. Similarly to [9], we use PCA [13] on raw descriptors to remove dimensions with low signal-to-noise ratio.

*D. Approximate Nearest Neighbor Search*

In contrast to most place recognition pipelines that use a BoW model, we directly perform kNN search on projected descriptors. To achieve this, a database of descriptors that describe visited places has to be built. Lynen *et al.* [10] used a k-d tree [14] to find nearest neighbors of query descriptors. Unfortunately, it is not trivial to add descriptors to the tree after it has been constructed because it can become unbalanced, impairing search performance. As a result, k-d trees are unsuitable for maps that change with time. A valid alternative for dynamic maps, which would be the case for an algorithm running online, is the inverted multi-index [15]. An extensive evaluation and justification for using it as part of a localization system was performed in [16]. The conclusion was that querying a projected descriptor in an inverted multi-index is multiple times faster compared to using a k-d tree, while at the same time offering comparable performance. For that reason, evaluation was performed using an inverted multi-index of projected descriptors.

*E. Vote Aggregation*

After retrieving the nearest neighbor of a query descriptor, our algorithm increases the vote count of the vertex to which the nearest neighbor descriptor belongs by one. The aggregation procedure is outlined in more detail as part of algorithm 1 on page 3. The reason for introducing the heuristic time delay in line 1 is the prevention of retrieving descriptor matches to places too close in time.

*F. Probabilistic Scoring*

After vote aggregation, each vertex has to receive a score that can be used to evaluate similarity to the query vertex. The naive approach would be counting the number of votes and applying heuristic normalization, as done in [9] for example. In most cases, threshold parameters using the number of votes are not intuitive and vary depending on the environment. This issue is addressed by formulating the problem in a way that allows for a probabilistic interpretation.

---

[1]Database refers to the set of descriptors or groups thereof that have been stored to represent places.

## Algorithm 1 Vertex-to-Vertex Loop Closure Detection

```
 1: t_delay ← 10 seconds                  ▷ Time delay of the database
 2: t_q ← 0                               ▷ Query time
 3: n_Database ← 0                        ▷ Number of vertices associated with the database
 4: D_Database ← ∅                        ▷ Set of descriptors in the database
 5: loop
 6:     % Vote Aggregation Procedure:
 7:     UPDATE(t_q)
 8:     v_q := v(t_q)                     ▷ vertex at query time t_q
 9:     t_Database ← t_q − t_delay
10:     if t_Database ≥ 0 then
11:         D_Database ← {D_Database ∪ PROJECTDESCRIPTORS(v(t_Database))}
12:         n_Database ← n_Database + 1
13:     end if
14:     for all k ∈ {[0, n_Database] ∩ N_>0} do
15:         c_k ← 0                       ▷ number of votes for k-th vertex v_k
16:     end for
17:     D_q ← PROJECTDESCRIPTORS(v_q))
18:     for all d ∈ D_q do
19:         SET(k_NN)                     ▷ according to table II
20:         for all j ∈ {[1, k_NN] ∩ N} do
21:             d_NN := j-th nearest neighbor of d in D_Database
22:             v_NN := vertex to which d_NN belongs
23:             c_NN ← c_NN + 1           ▷ increase vote count of v_NN
24:         end for
25:     end for
26:     % Loop Closure Detection:
27:     LOOPDETECTION({c_1, ..., c_n_Database})  ▷ see algorithm 2, page 4
28: end loop

29: function PROJECTDESCRIPTORS(v)
30:     D ← ∅
31:     for all descriptors d in vertex v do
32:         D ← {D ∪ PROJECTION(d)}
33:     end for
34:     return D
35: end function
```

The derivation of the probabilistic score is based on the assumption that, in case of exploring previously unknown places, each vote corresponds to a *random* descriptor in the database. Given this assumption, the number of matches for each vertex in the database is a binomial distribution. Let $X_i(t)$ be the random variable for the number of aggregated votes of vertex $i$ at time $t$, then

$$X_i(t) \sim \text{Bin}(n, p), \quad n = N(t), \quad p = \frac{\gamma_i}{\Gamma(t)}. \quad (1)$$

TABLE I
PARAMETER LIST: BINOMIAL DISTRIBUTION

| | |
|---|---|
| $x_i(t)$ | : Number of votes for vertex $i$ at time $t$ |
| $N(t) := \sum_i x_i(t)$ | : Total number of votes at time $t$ |
| $\gamma_i$ | : Number of descriptors in vertex $i$ |
| $\Gamma(t) := \sum_i \gamma_i$ | : Number of descriptors in inverted multi-index |

The parameter list is presented in table I. We anticipate that the number of votes for a vertex in the database will not follow the binomial distribution model if it represents the same place as the query vertex. Hence, we formulate the null hypothesis $H_0$:

*The number of matches $x_i(t)$ is drawn from a binomial distribution.*

As visualized in figure 2, $H_0$ is rejected if

$$\Pr(X_i(t) = x_i(t)) < \alpha < 1. \quad (2)$$

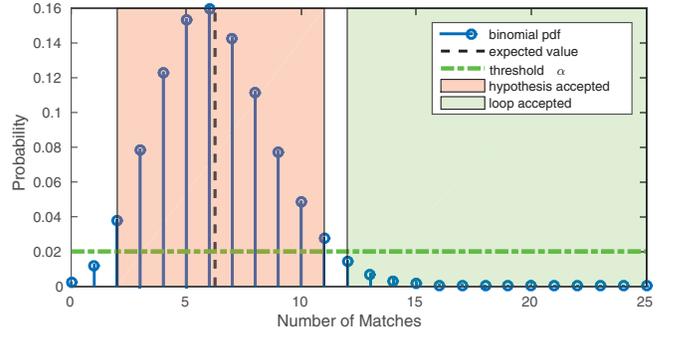

Fig. 2. A loop is only accepted if the number of matches is unlikely to be sampled from the computed binomial distribution.

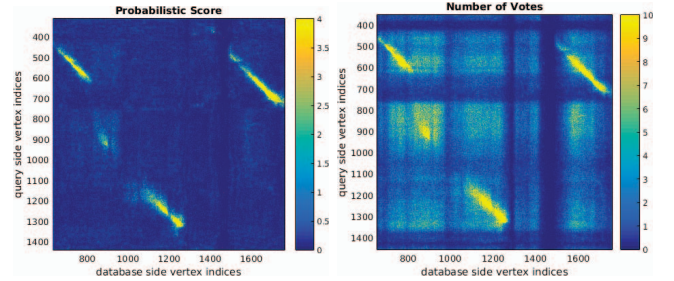

Fig. 3. Comparison between counting votes and probabilistic scoring on a dataset recorded by a fixed-wing aircraft. The plotted probabilistic score for each vertex is $-\log_{10}(\Pr(X_i(t) = x_i(t)))$ which indicates the implausibility that $x_i(t)$ is sampled from a binomial distribution. For illustration purposes, the probabilistic scores are limited to 4.

A loop with vertex $i$ is temporarily accepted if (2) holds and

$$x_i(t) > \mathrm{E}[X_i(t)] = N(t) \cdot \frac{\gamma_i}{\Gamma(t)}. \quad (3)$$

$1 - \alpha$ is the confidence required to accept a loop closure candidate vertex. Therefore, $1 - \Pr(X_i(t) = x_i(t))$ will be referred to as the *probabilistic score* of the $i$-th vertex. If (2) were the only check that a candidate vertex has to pass, we would not account for the case of having very few votes per vertex. Of course, this would indicate that the loop closure candidate should be rejected. Therefore, with condition (3), all loop closure candidates are discarded that have fewer votes than expected by random voting.

Note that this score is independent of the number of

- descriptors in the database,
- descriptors in the matched vertex,
- nearest neighbors returned for a given query vertex.

Consequently, there is no bias towards vertices with a large number of descriptors. In addition, this score is well suited for an online implementation since the size of the database is increasing with time. The effect of the probabilistic scoring is visualized by figure 3, for which the query and database vertex indices are ordered in increasing time. It is evident that the probabilistic formulation removes bias towards frames with a high number of features.

We now split the final part of the methodology to distinguish between two loop closure detector methods that match a query vertex to either a database vertex or a set of landmarks. Henceforth, they are labeled as 'vertex-to-vertex' and 'vertex-to-map'.

*1) Vertex-to-Vertex:* After probabilistic scoring of all vertices in the database, the vertex with the highest score is passed to geometric verification if conditions (2) and (3) are fulfilled. The scoring procedure is presented by algorithm 2. The complete loop closure detection algorithm including vote aggregation is shown in algorithm 1.

---

**Algorithm 2** Vertex-to-Vertex Probabilistic Scoring
---
1: **function** LOOPDETECTION($\{x_1, \ldots, x_{n_{\text{Database}}}\}$)
2:     Use constant threshold $\alpha \in (0, 1)$ as defined in equation (2)
3:     Use knowledge about $\{\gamma_1, \ldots, \gamma_{n_{\text{Database}}}\}$ as defined in table I
4:     $N_{\text{votes}} \leftarrow \sum_i^{n_{\text{Database}}} x_i$
5:     $\Gamma \leftarrow \sum_i^{n_{\text{Database}}} \gamma_i$
6:     $P_{\min} \leftarrow 1$
7:     **for all** $i \in \{[0, n_{\text{Database}}] \cap \mathbb{N}_{>0}\}$ **do**
8:         $P_i \leftarrow \text{Bin}(N_{\text{votes}}, \frac{\gamma_i}{\Gamma})$
9:         **if** $P_i < \min\{\alpha, P_{\min}\}$ and $x_i > N_{\text{votes}} \cdot \frac{\gamma_i}{\Gamma}$ **then**
10:             $P_{\min} \leftarrow P_i$
11:             $i_{\min} \leftarrow i$
12:         **end if**
13:     **end for**
14:     **if** $P_{\min} < 1$ **then**
15:         Perform geometric verification of the potential loop between the query vertex and $i_{\min}$-th vertex in the database
16:     **end if**
17: **end function**

---

*2) Vertex-to-Map:* With some adaptations, the algorithm can be applied to extract the set of landmarks that represent the queried place most accurately.

Instead of storing all descriptors that are tracked with the SLAM system, only the descriptors that are associated with a landmark are retained. As a consequence, each descriptor that is added to the inverted multi-index has an associated landmark. A landmark is usually observed by many vertices and is therefore linked by multiple descriptors.

Instead of just voting for the vertex that is associated with the descriptor's nearest neighbor, we vote for all vertices that not only observe the landmark but also lie within a certain timestep of the matched vertex. Assuming that the vertex containing the query descriptor's nearest neighbor has a timestamp $t_{\text{match}}$, we increase the vote count by one for all vertices with timestamp $t \in [t_{\text{match}} - \Delta t, t_{\text{match}} + \Delta t]$ that also observe the matched landmark.

Voting for all vertices that observe the matched landmark would introduce bias towards vertices that observe landmarks with long tracks. On the contrary, only voting for the vertex that contains the descriptor might not generate a sufficient number of votes to achieve high recall. This stems from the fact that usually less than 20% of the tracked features have an associated landmark that is useful for absolute pose estimation. For evaluation in section IV, $\Delta t$ was set to 1 second.

After computing the probabilistic score for each vertex in the database the following steps are performed:

a) Extract the set $\Psi_h$ that contains all vertices sharing at least one landmark observation with the vertex that received the highest score.

b) Compute $\Psi_\alpha \subseteq \Psi_h$, the set of vertices that have a score equal or larger than $1 - \alpha$ as defined in (2). It is also possible to choose a less restrictive threshold to compensate conservative choices of $\alpha$.

c) The set of landmarks that are observed by $\Psi_\alpha$ are passed to geometric verification.

This procedure is visualized by figure 4 which is a zoomed-in version of figure 1. Apart from the mentioned modifications, algorithm 1 and 2 also apply to the vertex-to-map case.

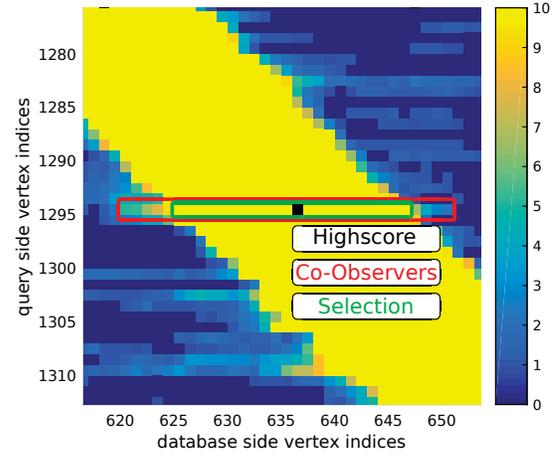

Fig. 4. Bright yellow squares symbolize loop closure candidate vertices. The black square represents the vertex in the database with the highest score. The red rectangle encompasses all vertices $\Psi_h$ that share at least one landmark observation with the black vertex. The green rectangle represents the sequence of vertices $\Psi_\alpha$ from which landmarks are extracted. The negative logarithmic probabilistic scores (as defined in figure 3) are limited to 10 for illustration purposes.

### G. Geometric Verification

Most place recognition pipelines eventually perform absolute or relative pose estimation to either localize in a given map or to ensure that the retrieved place is geometrically consistent with the query frame. Our loop closure detection algorithm is concerned with the latter case. Implementations of the following camera pose computation methods are open source and published in [17].

*1) Vertex-to-Vertex:* Given the query and matched vertex, our implementation builds a k-d tree with the projected descriptors of the matched vertex. Subsequently, two nearest neighbor descriptors are retrieved for each descriptor in the query image and a ratio test [18] is performed to remove outlier matches. The remaining matches are used to estimate the relative position of the query frame with respect to the matched frame using the 5-point method presented in [19] for a single camera setup and [20] for multi-camera systems in a RANSAC scheme [21].

*2) Vertex-to-Map:* We can directly perform absolute pose estimation on retrieved landmark matches. For single and multi-camera setups we use [22] in a preemptive RANSAC loop.

## III. IMPLEMENTATION DETAILS

*1) Approximation of Binomial Distribution:* Computing large numbers of probabilities from binomial distributions

can be expensive. Therefore we suggest approximating the binomial distribution with a poisson distribution under certain conditions. According to the Poisson limit theorem [23]; if

$$n \to \infty, \ p \to 0, \ \text{such that} \ np \to \lambda \quad (4)$$

then

$$\frac{n!}{(n-k)!k!}p^k(1-p)^{n-k} \to e^{-\lambda}\frac{\lambda^k}{k!}. \quad (5)$$

(1) or (5) have to be computed only if condition (3) is met. The approximation quality improves with decreasing $\lambda$ values. In most cases, $n$, the number of votes, is large enough to justify this simplification. In addition to that, with a growing pose-graph, $\lambda$ converges to 0 because $p$ converges to 0. At the same time, $n$ increases only slowly with the size of the database as explained in section III-.3. Consequently, the approximation quality improves over time.

Our implementation switches to the poisson approximation if $n \geq 200$ and $\lambda \leq 1$ for the vertex-to-vertex loop closure detector and $n \geq 2000$ and $\lambda \leq 20$ for the vertex-to-map variant of the algorithm. Different thresholds are used because the number of votes for the vertex-to-map method is much higher, leading to a better approximation even for larger values of $\lambda$. These thresholds were determined heuristically and could be improved by making use of statistical arguments. Usually, the approximation is valid after a few hundred vertices for the vertex-to-map method and a few thousand vertices for the vertex-to-vertex loop detector. We suggest using double-precision floating-point format for computing probabilities as well as pre-computing a look-up table of factorials to speed up computation.

*2) Thresholding Descriptor Distance:* Matching quality of both systems can be increased by employing a threshold on maximal descriptor distance. This parameter depends on the projection method, the number of dimensions in projected space, and utilized descriptor.

*3) Adaptive Number of Nearest Neighbors:* Our implementation also adapts the number of nearest neighbors that are retrieved for a query descriptor to the database size as illustrated in table II. This is done to compensate the decrease of nearest neighbor recall with growing database size.

TABLE II
NUMBER OF SEARCHED NEAREST NEIGHBORS $k_{NN}$ FOR A DESCRIPTOR.

| Database size | $k_{NN}$ |
|---|---|
| $< 1e4$ | 1 |
| $< 1e5$ | 2 |
| $< 1e6$ | 3 |
| $< 1e7$ | 6 |
| $\geq 1e7$ | 8 |

IV. EXPERIMENTAL VALIDATION

In order to evaluate our proposed systems, this section provides experiments on both public and in-house datasets that were recorded in vastly different environments. Along with describing the experimental setup, we provide an analysis and discussion of the relative performance of the methods.

*A. Test Sequences*

The algorithms are evaluated on four different test sequences: two urban driving sequences, one dynamic dataset from a multicopter flying in a machine hall, and one from a fixed-wing aircraft flying over a rural area. To give a notion of their diversity, groundtruth trajectories and example images are shown in figure 5. Two of the four sequences are from the KITTI Visual Odometry dataset collection [24]. Sequences 00 and 05 are selected because they provide relevant examples of loop closures in urban environments with accurate groundtruth, and are commonly used in evaluating place recognition. Furthermore, sequence 05 contains examples of perceptual aliasing as illustrated by figure 6. The EuRoC MH 05 difficult dataset [25] is selected to evaluate robustness against strong variations in velocity along the trajectory and strong perspective changes. In addition to these public datasets, we provide evaluations on an in-house aerial imagery dataset in order to analyze how the algorithms perform in an environment with strong perceptual aliasing, repetitive features, and illumination changes due to cloud coverage. It mainly contains fields and farm tracks as shown in figure 6.

*B. Experimental setup*

Our feature tracking pipeline is based on an ORB detector [26] that detects up to 2000 keypoints in each frame. Subsequently, weak keypoints are suppressed in an 8-pixel radius. The remaining features are tracked by a combination of matching descriptors and LK-tracking [27]. Moreover, we use the visual-inertial odometry framework proposed by Leutenegger *et al.* [28] to build a sparse map of landmarks. In the case of the KITTI datasets, only a single camera has been used in our evaluation, in order to facilitate parsing with our SLAM framework. Stereo cameras were used for detecting loops in the EuRoC dataset while the aerial dataset was recorded using only a single camera.

All settings, also those mentioned in previous sections, remain the same for all datasets.

*1) Precision and recall:* Groundtruth data is used to determine where loops exist in the data. In order to make use of this information, we define two distance thresholds between potential groundtruth matches as done in [10]. First, we define $d_{\text{near}}$ which is the maximal distance between poses to form a true match. Second, $d_{\text{far}}$ represents the minimal distance for classifying a match as false. We do not classify matches with distance $d \in (d_{\text{near}}, d_{\text{far}})$ as it remains unclear if the match is correct or not. Parameter settings for the public datasets are shown in table III.

*2) Baseline comparison:* We benchmark the proposed systems against a localization method suggested by Sattler *et al.* [29] using the same voting framework. Precision and recall are generated by varying the minimum number of 2D-3D matches. Since localization methods tend to have low precision by design, we also evaluate maximum recall at 100% precision *after* geometric verification.

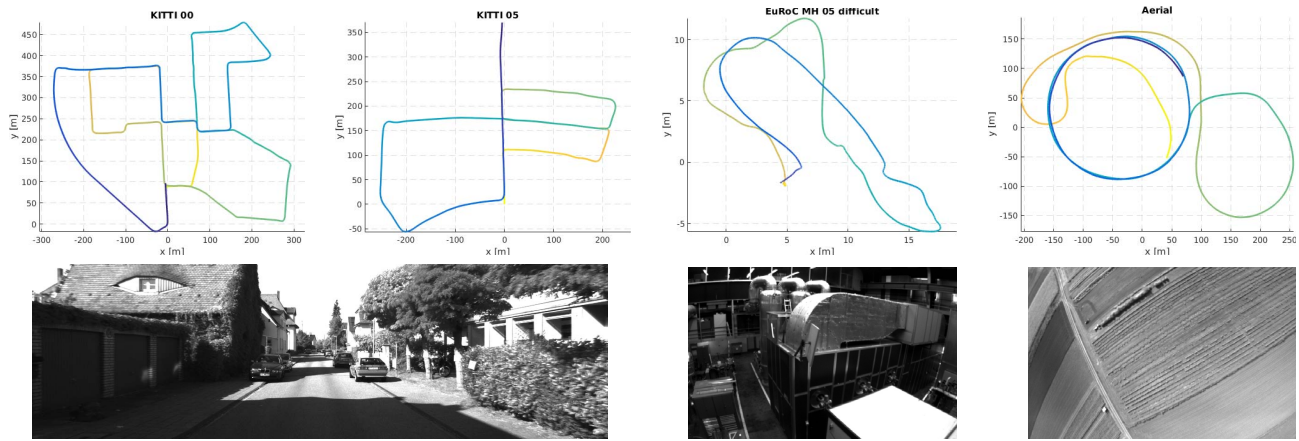

Fig. 5. Top down view on trajectories of all test sequences with example images. The EuRoC and Aerial trajectory also have varying altitude that is not visible in this figure.

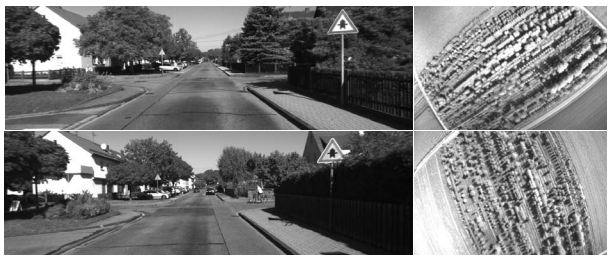

Fig. 6. Examples of perceptual aliasing and repetitive features: All images belong to different places, which implies that our algorithm should not match them. Left: KITTI 05. Right: Aerial.

TABLE III
PRECISION/RECALL PARAMETERS FOR ALL TEST SEQUENCES.

|  | $d_{\text{near}}$ [m] | $d_{\text{far}}$ [m] |
|---|---|---|
| KITTI 00/05 | 5 | 10 |
| EuRoC MH 05 difficult | 2 | 3.5 |
| Aerial | 20 | 25 |

*C. Results*

Figure 7 shows precision and recall plots for each dataset. Both methods described in section II-F reach over 90% recall at 100% precision on KITTI 00, while the performance of the vertex-to-map algorithm drops slightly in case of the KITTI 05 dataset. The cause for this slight deterioration in performance is likely due to the stronger examples of perceptual aliasing appearing in KITTI 05. Interestingly, the vertex-to-vertex algorithm's performance on both KITTI datasets is approximately equivalent with 99% precision at 95% recall. A possible explanation could be that this method approaches the best possible results with the employed definition of true and false matches based on distances. To the best of our knowledge, no other loop closure detection algorithm has surpassed this performance on these datasets. Moreover, the presented results are achieved with low-dimensional feature descriptors in order to maintain efficiency, and we expect results could be improved by the use of more information-rich features.

On the EuRoC dataset, both the vertex-to-vertex and vertex-to-map methods perform comparably. We surmise that the decisive factor influencing performance for this dataset is related to the ORB detector's limitations: it is not able to detect the same features if the scene is observed from a significantly different perspective. On the other hand, both of the proposed methods appear to be robust against the changes in velocity which occur in the EuRoC sequence.

The aerial dataset seems to be particularly challenging for the vertex-to-map loop detector. In fact, it is not possible to reach 100% precision with this method without an additional geometric verification step. For this dataset, the vertex-to-vertex algorithm clearly outperforms the others indicating enhanced robustness against strong perceptual aliasing

Generally speaking, the vertex-to-vertex algorithm outperforms the vertex-to-map method. We can offer two possible interpretations for this observation:

a) The vertex-to-map system only adds descriptors which belong to triangulated 3D landmarks to the inverted multi-index. Consequently, it has much less information to work with than the vertex-to-vertex algorithm, as many tracked features are not mapped to a corresponding landmark.

b) The parameter $\Delta t$ introduced in section II-F.2 can increase sensitivity to perceptual aliasing. For example, if multiple images close in time receive many false positive matches, they increase the number of votes of each other. The vertex-to-vertex method does not suffer from this to the same degree because it only votes for a single image. A possible solution would be aggregating votes of multiple vertices and scoring them together[2]. Different aggregation techniques are evaluated e.g. in [10].

We did not compare the baseline method to the proposed methods on the KITTI datasets because it achieved comparable performance after geometric verification. The post-RANSAC performance differences are more evident for the EuRoC and Aerial datasets for which the baseline method clearly underperforms. We therefore believe that the missing normalization over descriptors in vertices could lead to wrong connectivity in the co-visibility graph. Another interesting result is that the post-RANSAC recall of the vertex-to-vertex method is rather low in the EuRoC dataset. This indicates that relative pose estimation is more susceptible to strong perspective changes than absolute pose estimation.

---

[2]This also applies to the vertex-to-vertex algorithm.

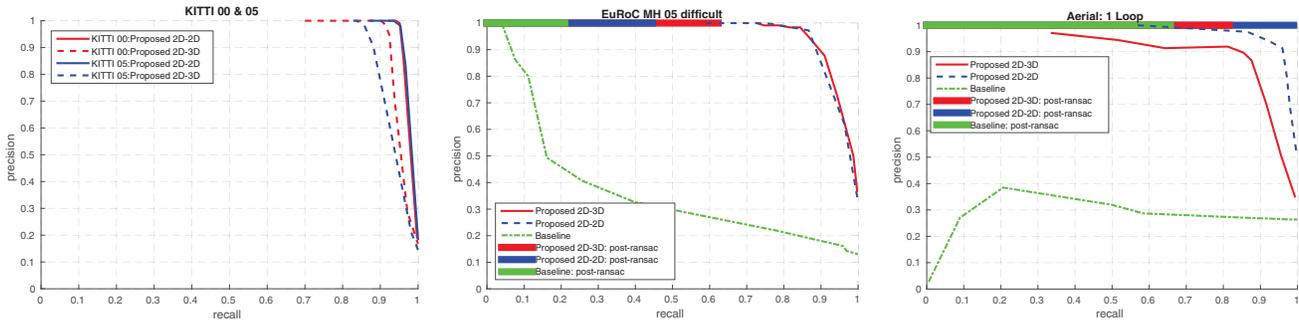

Fig. 7. Comparison of precision-recall results for the proposed methods on the 'KITTI 00', 'KITTI 05', 'EUROC MH 05' and 'Aerial' datasets. Color bars at the top of the plot indicate the achievable recall at full precision for each respective method after geometric verification.

The runtime for the methods proposed in section II-F on the KITTI 00 dataset are shown in table IV. Here, 'query' refers to a query of a vertex while 'add' stands for projection of all descriptors plus inserting them into the inverted multi-index. The difference in the query timing mainly stems from the fact that retrieval of $\Psi_\alpha$, introduced in section II-F.2, is not optimized in our code. As expected, the vertex-to-map method takes less time to insert descriptors of a vertex into the database because only few descriptors have a corresponding landmark. Total average runtime of all evaluated datasets is between 30 to 40 milliseconds per query vertex.

TABLE IV
RUNTIME (INTEL CORE I7 4700 MQ) ON KITTI 00 IN MILLISECONDS.

|  |  | *Vertex-to-Map* avg | std | *Vertex-To-Vertex* avg | std |
|---|---|---|---|---|---|
| KITTI 00 | query | 36.1 | 4.8 | 22.9 | 6.7 |
|  | add | 3.0 | 1.6 | 7.6 | 3.2 |
|  | total | **39.1** | - | **30.5** | - |

## V. CONCLUSION

In this paper, we have introduced a probabilistic approach to improve visual place recognition frameworks which are based on descriptor voting techniques. By considering the scenario of exploring unknown places, we derived a probabilistic score that is invariant to various parameters of the database and stored places. This score was subsequently used in conjunction with two loop detector methods that aggregate votes per vertex and demonstrated high performance over a range of different datasets. The resulting methods are additionally shown to be suitable for online, real-time operation. The proposed loop closure detection algorithms considered a single image or vertex as a place. However, related work such as [10] indicates that determining the correct notion of a place is crucial for place recognition. The score that was introduced with equation (1) can be defined for arbitrary place formulations by setting $\gamma_i$ to the number of descriptors in the current map that are associated with place $i$. We expect that combining probabilistic scoring together with more sophisticated definitions of places could enhance place recognition performance of voting schemes even more.

## REFERENCES

[1] S. Lowry, N. Sünderhauf, P. Newman, J. J. Leonard, D. Cox, P. Corke, and M. J. Milford, "Visual place recognition: A survey," *IEEE Trans. on Robotics*, vol. 32, no. 1, pp. 1–19, 2016.
[2] S. Thrun and M. Montemerlo, "The graph slam algorithm with applications to large-scale mapping of urban structures," *The Int. Journal of Robotics Research*, vol. 25, no. 5-6, pp. 403–429, SAGE Publications.
[3] J. Sivic and A. Zisserman, "Video google: A text retrieval approach to object matching in videos," in *Int. Conf. on Computer Vision*, 2003.
[4] M. Cummins and P. Newman, "Appearance-only slam at large scale with fab-map 2.0," *The Int. Journal of Robotics Research*, vol. 30, no. 9, pp. 1100–1123, 2011.
[5] D. Gálvez-López and J. D. Tardos, "Bags of binary words for fast place recognition in image sequences," *IEEE Trans. on Robotics*, vol. 28, no. 5, pp. 1188–1197, 2012.
[6] E. Stumm, C. Mei, S. Lacroix, J. Nieto, M. Hutter, and R. Siegwart, "Robust visual place recognition with graph kernels," in *IEEE Conf. on Computer Vision and Pattern Recognition*, 2016.
[7] H. Jégou, M. Douze, and C. Schmid, "Hamming embedding and weak geometric consistency for large scale image search," in *European Conf. on Computer Vision*, 2008.
[8] G. Schindler, M. Brown, and R. Szeliski, "City-scale location recognition," in *IEEE Conf. on Computer Vision and Pattern Recognition*, 2007.
[9] T. Cieslewski, E. Stumm, A. Gawel, M. Bosse, S. Lynen, and R. Siegwart, "Point cloud descriptors for place

recognition using sparse visual information," in *IEEE Int. Conf. on Robotics and Automation*, 2016.

[10] S. Lynen, M. Bosse, P. Furgale, and R. Siegwart, "Placeless place-recognition," in *Int. Conf. on 3D Vision*, vol. 1, 2014.

[11] G. Grisetti, R. Kummerle, C. Stachniss, and W. Burgard, "A tutorial on graph-based slam," *IEEE Intelligent Transportation Systems Magazine*, vol. 2, no. 4, pp. 31–43, 2010.

[12] S. Leutenegger, M. Chli, and R. Y. Siegwart, "Brisk: Binary robust invariant scalable keypoints," in *Int. Conf. on Computer Vision*, 2011.

[13] I. Jolliffe, *Principal component analysis*. Wiley Online Library, 2002.

[14] J. L. Bentley, "Multidimensional binary search trees used for associative searching," *Communications of the ACM*, vol. 18, no. 9, pp. 509–517, 1975.

[15] A. Babenko and V. Lempitsky, "The inverted multi-index," in *IEEE Conf. on Computer Vision and Pattern Recognition*, 2012.

[16] S. Lynen, T. Sattler, M. Bosse, J. Hesch, M. Pollefeys, and R. Siegwart, "Get out of my lab: Large-scale, real-time visual-inertial localization," in *Robotics: Science and Systems*, 2015.

[17] L. Kneip and P. Furgale, "Opengv: A unified and generalized approach to real-time calibrated geometric vision," in *IEEE Int. Conf. on Robotics and Automation*, 2014.

[18] D. G. Lowe, "Distinctive image features from scale-invariant keypoints," *Int. Journal of Computer Vision*, vol. 60, no. 2, pp. 91–110, 2004.

[19] H. Stewenius, C. Engels, and D. Nistér, "Recent developments on direct relative orientation," *ISPRS Journal of Photogrammetry and Remote Sensing*, vol. 60, no. 4, pp. 284–294, 2006.

[20] L. Kneip and H. Li, "Efficient computation of relative pose for multi-camera systems," in *IEEE Conf. on Computer Vision and Pattern Recognition*, 2014.

[21] M. A. Fischler and R. C. Bolles, "Random sample consensus: a paradigm for model fitting with applications to image analysis and automated cartography," *Communications of the ACM*, vol. 24, no. 6, pp. 381–395, 1981.

[22] L. Kneip, P. Furgale, and R. Siegwart, "Using multi-camera systems in robotics: Efficient solutions to the npnp problem," in *IEEE Int. Conf. on Robotics and Automation*, 2013.

[23] A. Papoulis and S. U. Pillai, *Probability, random variables, and stochastic processes*. Tata McGraw-Hill Education, 2002.

[24] A. Geiger, P. Lenz, C. Stiller, and R. Urtasun, "Vision meets robotics: The kitti dataset," *The Int. Journal of Robotics Research*, p. 0278364913491297, 2013.

[25] M. Burri, J. Nikolic, P. Gohl, T. Schneider, J. Rehder, S. Omari, M. W. Achtelik, and R. Siegwart, "The euroc micro aerial vehicle datasets," *The Int. Journal of Robotics Research*, vol. 35, no. 10, pp. 1157–1163, 2016.

[26] E. Rublee, V. Rabaud, K. Konolige, and G. Bradski, "Orb: An efficient alternative to sift or surf," in *Int. Conf. on Computer Vision*, 2011.

[27] B. D. Lucas and T. Kanade, "An iterative image registration technique with an application to stereo vision." in *Int. Joint Conf. on Artificial Intelligence*, vol. 81, no. 1, 1981, pp. 674–679.

[28] S. Leutenegger, S. Lynen, M. Bosse, R. Siegwart, and P. Furgale, "Keyframe-based visual–inertial odometry using nonlinear optimization," *The Int. Journal of Robotics Research*, vol. 34, no. 3, pp. 314–334, 2015.

[29] T. Sattler, B. Leibe, and L. Kobbelt, "Improving image-based localization by active correspondence search," in *European Conf. on Computer Vision*. Springer, 2012.